\documentclass[journal]{IEEEtranTIE}
\pdfoutput=1
\usepackage{cite}
\usepackage{amsmath,amssymb,amsfonts}
\usepackage{algorithm}
\usepackage{algorithmic}
\usepackage{graphicx}
\usepackage{epstopdf}
\usepackage{textcomp}
\usepackage{array}
\usepackage{hyperref}
\usepackage{multirow}
\usepackage{color}
\usepackage{bm}
\usepackage{subfigure}
\usepackage{units}
\usepackage{siunitx}
\usepackage{makecell}
\usepackage{balance}
\definecolor{gray}{RGB}{150,150,150}

\begin{document}
\title{MITNet: GAN Enhanced Magnetic Induction Tomography Based on Complex CNN}
\author{
	\vskip 1em
	Zuohui~Chen,
	Qing~Yuan,
    Xujie~Song,
    Cheng~Chen,
    Dan~Zhang,
    Yun~Xiang$^{\dag}$,
    Ruigang~Liu$^{\dag}$,
    and Qi~Xuan

	\thanks{

    This work was supported by the National Natural Science Foundation of China (61572439, 61771476) and Zhejiang Provincial Natural Science Foundation of China (LY18F030021, LR19F030001).
    Y. Xiang and R. Liu are the co-corresponding authors of this work.
	Z. Chen, Q. Yuan, X. Song, and Y. Xiang (e-mail: xiangyun@zjut.edu.cn) are with the Institute of Cyberspace Security, Zhejiang University of Technology, Hangzhou 310023, China.
    D. Zhang is with the College of Information Engineering, Zhejiang University of Technology, Hangzhou 310023, China.
    R. Liu (e-mail: ruigang@fmmu.edu.cn) is with the Department of Medical Electronic Engineering, School of Biomedical Engineering, Fourth Military Medical University, Xi'an, China.
	C. Chen, and Q. Xuan is with the Utron Technology Co., Ltd., Hangzhou 310023, China.
}
	}

\maketitle
	
\begin{abstract}
Magnetic induction tomography (MIT) is an efficient solution for long-term brain disease monitoring, which focuses on reconstructing bio-impedance distribution inside the human brain using non-intrusive electromagnetic fields. However, high-quality brain image reconstruction remains challenging since reconstructing images from the measured weak signals is a highly non-linear and ill-conditioned problem. In this work, we propose a generative adversarial network (GAN) enhanced MIT technique, named MITNet, based on a complex convolutional neural network (CNN). The experimental results on real-world dataset validate the performance of our technique, which outperforms the state-of-art method by 25.27\%.
\end{abstract}


\markboth{}%
{}

\definecolor{limegreen}{rgb}{0.2, 0.8, 0.2}
\definecolor{forestgreen}{rgb}{0.13, 0.55, 0.13}
\definecolor{greenhtml}{rgb}{0.0, 0.5, 0.0}

\section{Introduction}

\label{sec:introduction}

\IEEEPARstart{C}{erebrovascular} diseases seriously threaten human health. They can cause diseases such as half-body dysfunction, speech disorders, and even death, etc. Thus, early prevention and treatment are essential. Existing diagnosis methods include computed tomography (CT) and magnetic resonance imaging (MRI), which are expensive, time-consuming, and radiation hazardous. An alternative and non-intrusive method is MIT, which is also known as electromagnetic tomography or mutual inductance tomography. It applies a harmless magnetic field to induce eddy currents in human brain tissues and then uses sensing coils to detect the magnetic field strengths. The magnetic field can penetrate the brain skull and detect the electrical conductivity of brain tissues. Therefore, in this work, we propose a deep learning enhanced MIT technique based on complex CNN and GAN.

Compared with the existing techniques, MIT is low-cost, harmless, and hence, more suitable technique for long-term and continuous monitoring. The brain conductivity map reconstruction process, which is also known as the MIT inverse problem, is a major challenge. Various linear algorithms, e.g., back-projection, Tikhonov regularization~\cite{wei2012four}, truncated singular value decomposition~\cite{shi2013greedy}, and Landweber iteration method~\cite{wei2012volumetric} etc., are proposed to solve the problem. However, since the MIT system is inherently nonlinear and ill-conditioned, these methods are generally ineffective~\cite{soleimani2006absolute}. They linearize and approximate the nonlinear problems, which can cause significant reconstruction error~\cite{chen2019a}.

Recently, deep learning based techniques are widely used in the medical imaging area. They include segmenting interested regions~\cite{rachmadi2018segmentation}, classifying cancer types~\cite{murtaza2019deep}, disease diagnosis~\cite{liu2019comparison}, and image reconstruction~\cite{yang2017dagan} etc. The deep learning based techniques have powerful learning and generalization abilities~\cite{xuan2017automatic,chen2020signet,xuan2019multiview}. However, the existing methods cannot be applied directly to reconstruct the impedance distribution map from measured signals.

Therefore, in this work, we develop an MIT based system and propose a complex CNN and GAN based algorithm to predict the object location and shape simultaneously. Specifically, we make the following contributions.
\begin{enumerate}
\item We collect a realistic dataset using our MIT system. This dataset is useful and important since most previous studies validate their methods only using simulated data, which can be inaccurate and misleading.
\item We propose a novel GAN enhanced impedance reconstruction algorithm based on complex CNN, i.e., MITNet. This method can be directly applied on complex-valued data without rounding off the data.
\item We evaluate our method on the real-world dataset. The experimental results validate that our algorithm outperforms the state-of-the-art ones significantly.
\end{enumerate}

The rest of paper is organized as follows. \autoref{sec:related work} presents the related work. \autoref{sec:Mechanism and Dataset} introduces the mechanism and details of our system and dataset. \autoref{sec:method description} describes the MITNet. \autoref{sec:experiments} gives the experimental setup and results. \autoref{sec:conclusion} concludes the paper.

\section{Related Work}
\label{sec:related work}
In this section, we generalize the related work into three categories, including magnetic induction tomography, deep complex networks, and generative adversarial networks, respectively.

\subsection{Magnetic Induction Tomography}

Recently, algorithms using deep learning and neural networks are becoming popular to solve the MIT problem. Xiang et al.~\cite{xiang2020multi} propose a multi-frequency electromagnetic tomography (mfEMT) method for the initial diagnosis of acute stroke. They use a sensor array consisting of 12 gradiometer coils with multi-frequency sine waves to excite and sense the sensing region. After a sequence of measurements is acquired, frequency-constrained sparse Bayesian learning is used to derive the distribution. Their method performs well in numerical simulation but the image quality went down on the phantom experiment. Chen et al.~\cite{chen2019a} use a stacked auto-encoder (SAE) neural network composed of a multi-layer automatic encoder to reconstruct the distribution of electrical characteristics. Their method assumes that the inner electrical characteristic change of the object can cause the change of phase difference in measurement, which is proportional to the conductivity. The phase difference values are calculated through the forward process with given conductivity distribution and then used to train the neural network. Their experiment is fully conducted on simulation data. Li et al.~\cite{li2018deepnis} find that there is a fundamental connection between a deep neural network (DNN) architecture and an iterative method of the MIT problem. Inspired by this connection, they propose a DNN architecture DeepNIS, which is a complex-valued residual CNN cascaded by multiple layers. They use the MNIST dataset to simulate numbers placed in the sensing region. A full-wave solver to Maxwell‘s equations is used to derive images. Essentially, they train an image translation network to polish the coarse output of a conventional approach. Moreover, their method is still validated only on simulation data.

The existing researches mostly use only simulation data to train and verify their methods, while the majority of which are traditional machine learning algorithms.
\subsection{Deep Complex Networks}

The sensing coils in our system can detect the signal amplitude and phase with the excitation signal. In other words, the measured data are complex-valued, which has the potential of rich representational capacity~\cite{arjovsky2016unitary}.

Complex-valued data is widely used in the field of audio processing and signal processing. Arjovsky et al.~\cite{arjovsky2016unitary} propose unitary evolution recurrent neural networks (Unitary Evolution RNNs) to solve the problem of long-term dependencies. They use orthogonal and unitary matrices in RNNs as building blocks. Their model uses complex-valued matrices and parameters and can provide a richer representation in the complex domain. Danihelka et al.~\cite{danihelka2016associative} use holographic reduced representations (HRRs) to model complex vectors using real vectors. They propose associative long short-term memory (LSTM) networks by combining LSTM with HRRs. They show that using complex-valued vectors is numerically more stable and efficient than real-valued matrices.

\subsection{Generative Adversarial Networks}

GAN typically consists of two networks: a generation network which generates output similar to the distribution of training data, and a discrimination network which scores the generated output. GAN is an unsupervised learning model, where the generator automatically learns the distribution with unlabeled data. Through the two-player game, we derive a generator undistinguishable from real samples.

GAN is first proposed by Goodfellow et al.~\cite{goodfellow2014generative}. The prototype leverage the adversarial relationship between generator and discriminator. However, the training of GAN is still challenging~\cite{hong2017generative}. Radford et al.~\cite{radford2015unsupervised} further improve the original design, and proposes the deep convolutional generative adversarial networks (DCGAN), which uses batch normalization~\cite{ioffe2015batch}, transposed convolution, and leaky ReLu~\cite{xu2015empirical}. Mirza and Simon~\cite{mirza2014conditional} introduce supervised learning into the training of GAN and propose conditional generative adversarial networks (CGAN) to enhance the controllability and limit the extent of output.

In medical image analysis, GAN provides a new solution option to many challenging problems like medical image de-noising, reconstruction, and data simulation~\cite{kazeminia2018gans}. Mandija et al.~\cite{mandija2019opening} model the electrical properties reconstruction problem in MRI as a supervised deep learning task. They employ two CGANs to generate binary masks and MRI images, respectively. The mask distinguishes tissue from the air while the image provides tissue contrast information. Frid et al.~\cite{frid2018gan} use GANs to generate synthesis medical images that augment data for CNN training. Pradhan et al.~\cite{pradhan2020transforming} present a GAN based method to transform 2-dimensional medical images into multiple dimensions.

\section{MIT System and Dataset}
\label{sec:Mechanism and Dataset}
In this section, we introduce the system design and the corresponding dataset in detail.

\subsection{MIT Measurement System}

The MIT system contains sets of excitation coils and sensing coils placed around the object. The excitation coils are energized with alternating current, which can induce eddy currents and the corresponding magnetic fields. The conductivity and dielectric constant distribution of the object can alter the intensity of eddy currents~\cite{xiao2018multi-frequency}. Therefore, based on the induced voltage captured by the sensing coils, we can reconstruct the object image.


\begin{figure} [!t]
\centering
\includegraphics[width=0.45\textwidth]{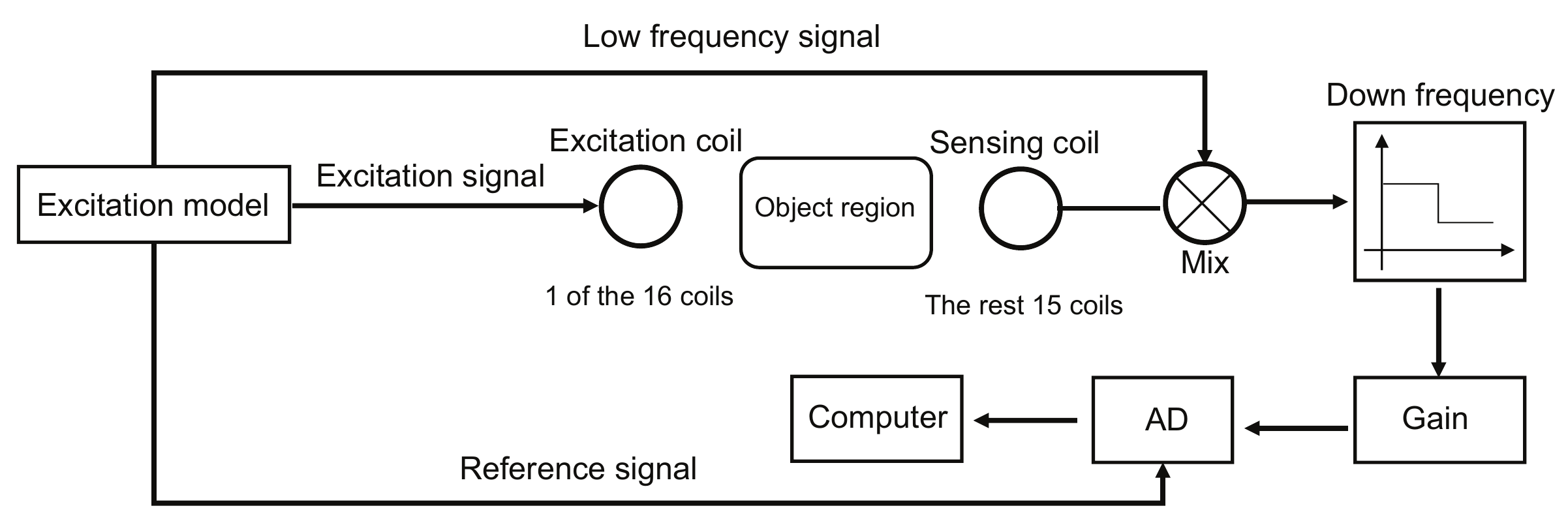}
\caption{Measurement diagram.}
\label{Fig:measure}
\end{figure}

The data acquisition system is completed in cooperation with Hangzhou Utron Technologies Co. Ltd~\cite{chen2021real}. The system includes four parts: a multiplexed coil array, signal generation and control module, signal detection unit, and an automatic gain module. The object field is a circle with a diameter of \SI{20}{\centi\meter} with 16 coils distributed at equal angles on the periphery. The coil array consists of 16 multiplexed coils, which have the same configuration, all with 13 turns and the coil diameter is \SI{18}{\milli\meter}. As shown in Fig.~\ref{Fig:measure}, in a round of measurement, each coil, in turn, excites the magnetic field while the rest receive the magnetic field interfered by the object. Since the system requires a signal source capable of outputting 16 excitation signals with very stable frequency and phase, FPGA and DAC modules are used here. Direct measurement of the phase of high-frequency signals can be difficult. By mixing it with a low frequency signal, a low-frequency signal containing high-frequency signal phase information is derived. The automatic gain module amplifies detected signals to achieve high-precision phase discrimination, then the AD module converts them to digital data.

\begin{figure} [!t]
\centering
\includegraphics[width=0.45\textwidth]{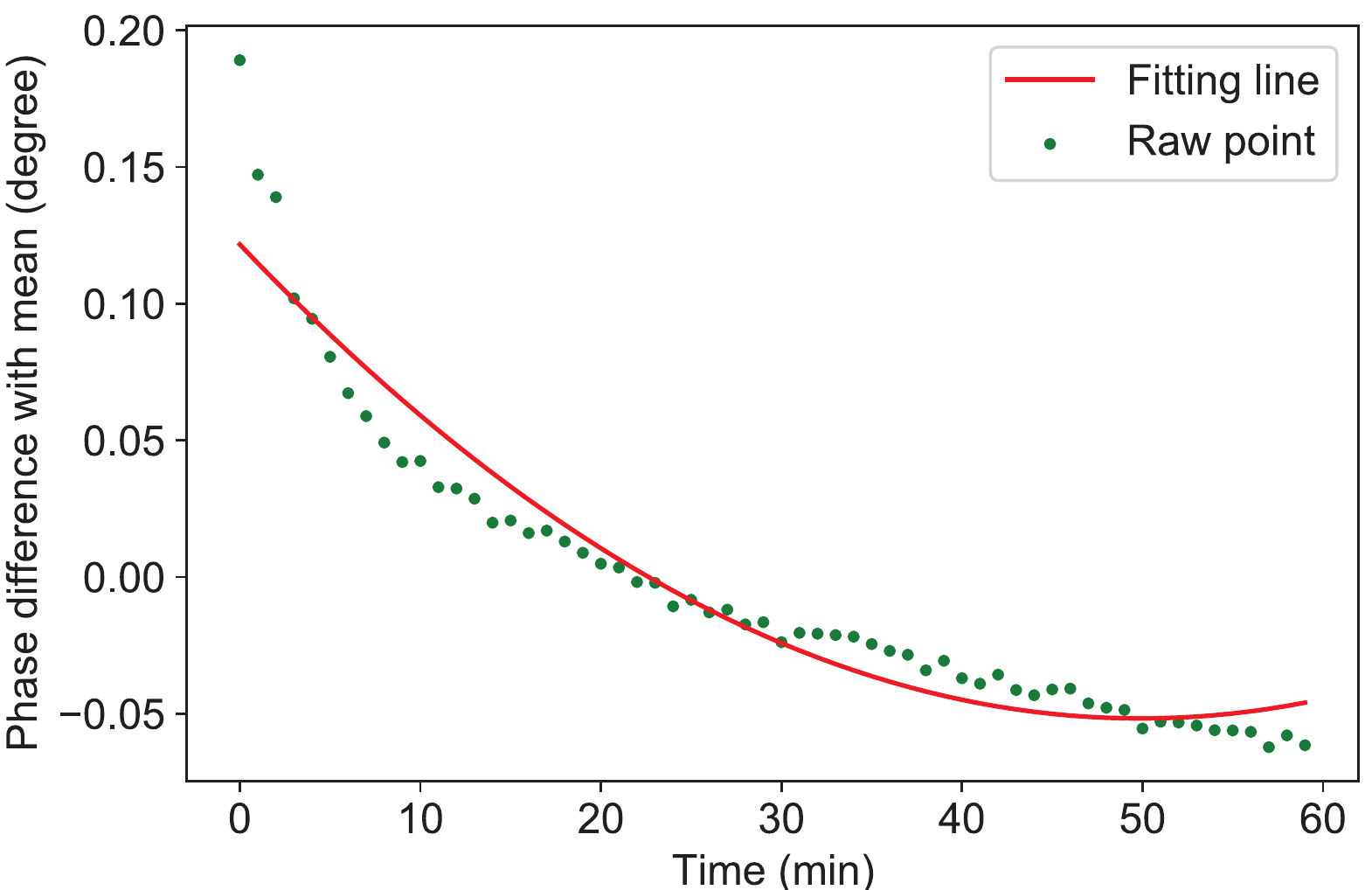}
\caption{The difference between the measured phase and its mean of one pair coils.}
\label{Fig:drift_one}
\end{figure}

The key metrics of an effective MIT system are sensitivity and stability. Thus, we evaluate the system performance, including signal noise ratio (SNR) and phase drift~\cite{chen2021real}. Since the system contains 16 multiplexed coils, we measure the SNR of every possible combination using the following equation.
\begin{equation}
SNR = 10\log_{10}{\frac{m^2}{v}},
\end{equation}
where $m$ is the mean and $v$ is the variance of the measured data. There is a total of 240 combinations of emitter pairs and the average SNR is \SI{62}{\dB}, the maximum is \SI{71}{\dB} and the minimum is \SI{51}{\dB}. Fig.~\ref{Fig:drift_one} shows the mean phase variation for one hour after the start-up. The measured data phase is becoming stable after \SI{40}{\minute} with an average variation of less than 0.05 degree.

\subsection{The MIT Dataset}

\begin{figure} [!t]
\centering
\includegraphics[width=0.5\textwidth]{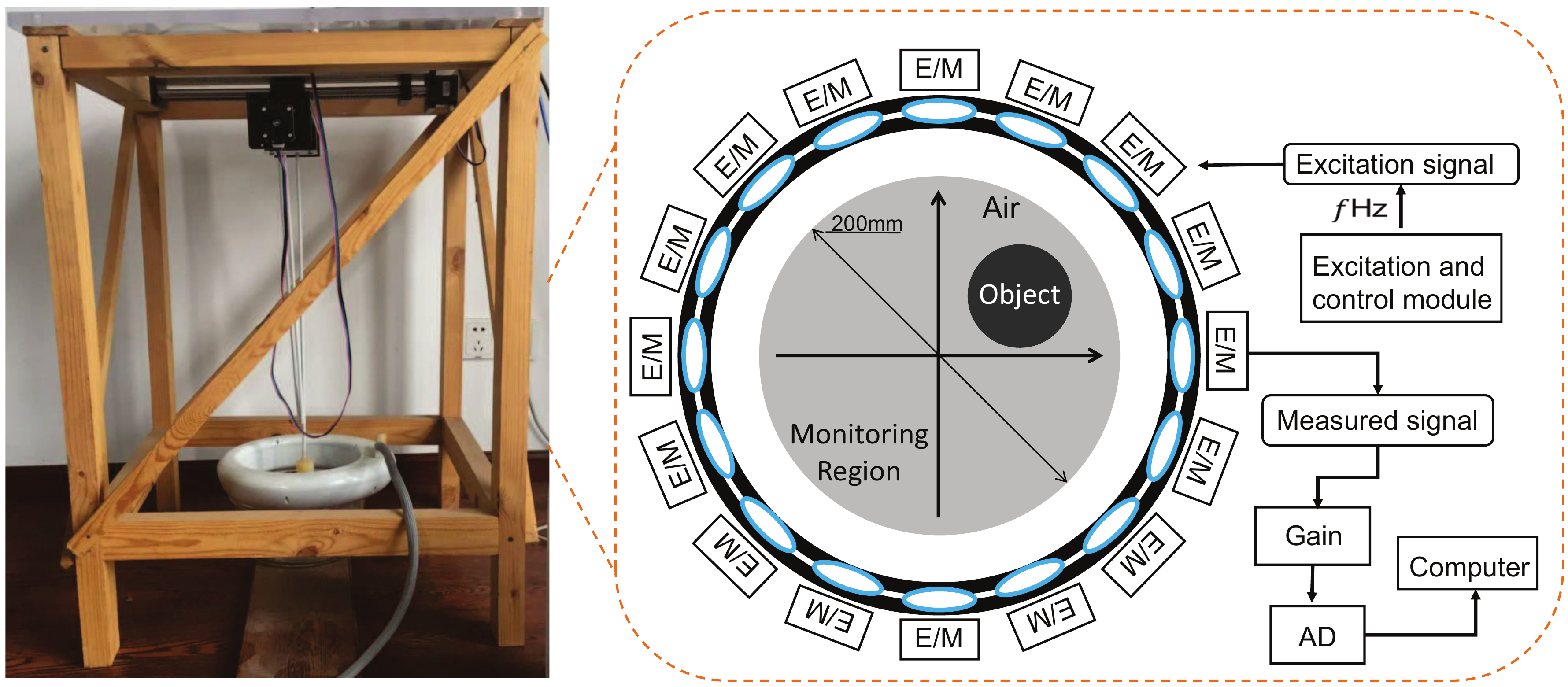}
\caption{MIT data acquisition system.}
\label{Fig:data-device}
\end{figure}

Most existing MIT researches only use simulation data, which can be quite different from real-world applications. In this work, we use our MIT system to collect a real-world dataset.

The data collection device we used is shown in \autoref{Fig:data-device}. In particular, a plastic cylinder filled with salt water is fixed on a screw slide rail using three rigid sticks. The rail is fixed on another screw slide rail which is perpendicular to the first one. The center of the imaging field is set as the origin point. Stepper motors are used to accurately move and locate the object. There are two different cylinders with diameters of \SI{30}{\milli\metre} (CY-3) and \SI{35}{\milli\metre} (CY-3.5), and a triangular prism (PR), as presented in TABLE~\ref{Tab:data_all}. The heights of all the objects are \SI{8}{\centi\metre}.

The cylinders are moved at a step of \SI{4}{\milli\metre} (Distance in the table), while the triangular prism is moved at a step of \SI{5}{\milli\metre}. In total, the cylinders cover 1,229 positions and the triangular prism covers 749 positions. The objects stay 20 frames (one frame means one round of measurement of all sensing coils) at each position to gain stable measurement, specifically we take the tenth frame as one position data. To get enough data, the above operations are repeated for three or four rounds (Rounds in the table). Note that the data collected under different batches are affected by the environment noise and mechanical instability, and thus are different. As far as we know, this is the first real-world MIT dataset large enough for deep learning training and evaluation.

\begin{table}[!t]
	\caption{Data of Different Objects}
	\label{Tab:data_all}
    \centering
	\begin{tabular}{|c|>{\centering}p{1.4cm}|>{\centering}p{0.8cm}|>{\centering}p{0.8cm}|>{\centering}p{0.8cm}|>{\centering}p{0.8cm}|}
		\hline
		\multirow{2}*{Object} & Conductivity  & Distance  & \multirow{2}*{Rounds} & Total \tabularnewline
        ~            & (\SI{}{\siemens\per\metre}) & (\SI{}{\milli\metre}) & ~    &  amount \tabularnewline
        \hline
		CY-3   & 2 & 4 & 4 & 3687\tabularnewline
	    CY-3.5 & 3 & 4 & 3 & 3687\tabularnewline
		PR     & 2 & 5 & 3 & 2247\tabularnewline
		\hline
	\end{tabular}
\end{table}

This dataset contains 3,687 samples for \SI{30}{\milli\metre} cylinder and \SI{35}{\milli\metre} cylinder each, and 2,247 samples for triangular prism. There are 1,229 positions and 749 positions for cylinders and prism, respectively. Each position has 3 measurements.

\section{MITNET: MIT Processing Network}
\label{sec:method description}

\begin{figure*} [!t]
\centering
\includegraphics[width=1.0\textwidth]{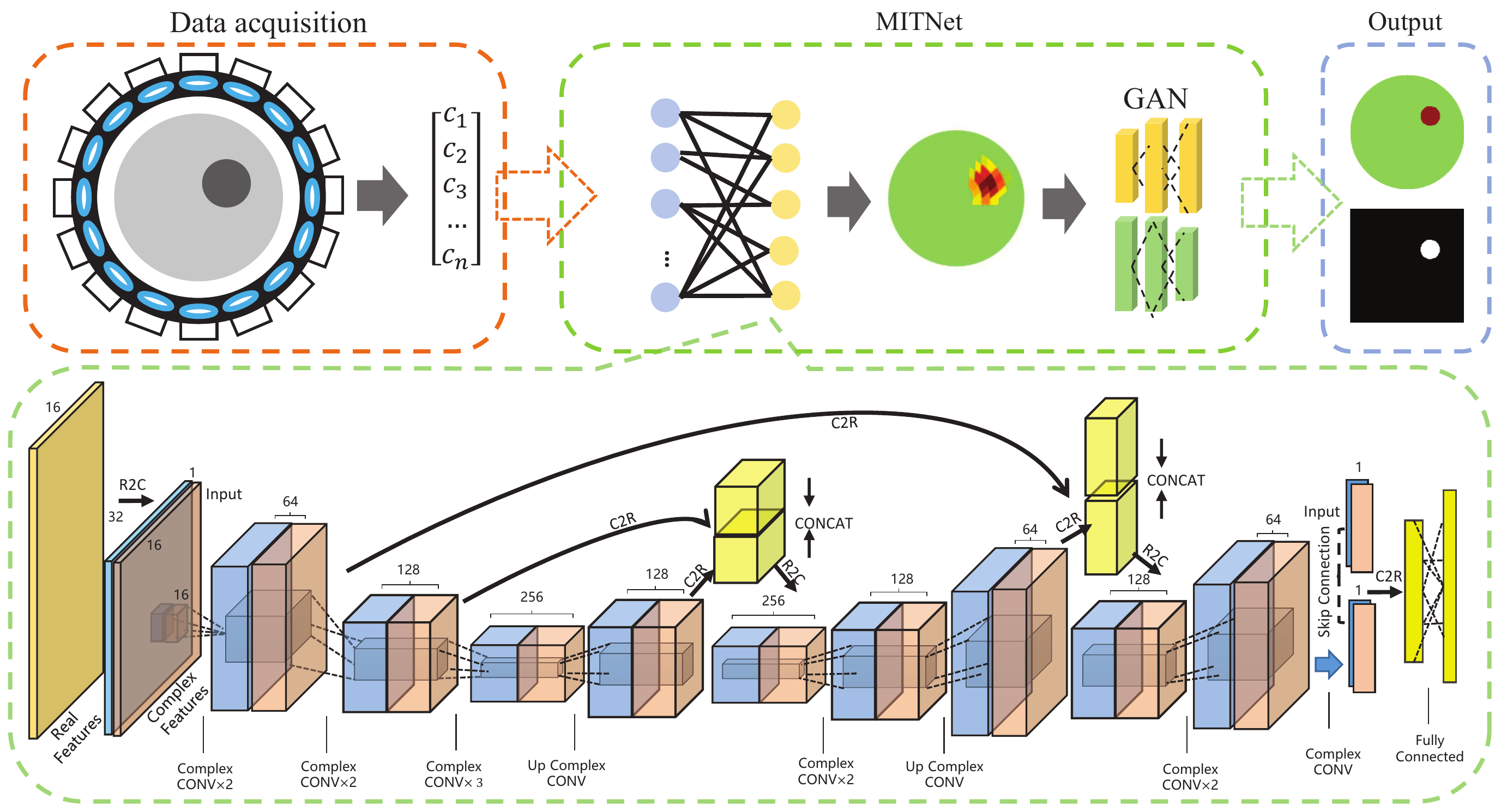}
\caption{The overall architecture of our system. }
\label{Fig:method-overall}
\end{figure*}

In this section, we introduce the structure of our system and the MITNet.

\subsection{MITNet Overview}

Fig.~\ref{Fig:method-overall} shows the overall architecture of the image reconstruction system. The data acquisition equipment collects the excitation signals and converts them into digital data, which is then fed into the MITNet deep learning structure. Specifically, the MITNet consists of two components: complex CNN and GAN. The complex CNN has a $16\times16$ complex matrix (differential frame) as input and reconstructs the locations and shapes of the objects. The GAN is used to enhance the reconstruction results. The complex CNN model utilizes both down-sampling and up-sampling structures. On the down-sampling side, between the complex convolution layers , there are several max-pooling layers. The input data is a $16 \times 32$ real-valued vector, which is then converted to a $16\times 16\times 2$ complex-valued input to the network. On the up-sampling side, to generalize and process the features from different layers, outputs from different layers are converted to real numbers (C2R) for concatenation and then to complex values (R2C) for convolution.

\subsection{Complex CNN}

A complex-valued DNN has unique building components~\cite{trabelsi2018deep}, including complex number representation, complex convolution, and complex-valued activations. But the current deep learning framework can only compute real numbers. We need to represent complex-valued tensor and implement complex number operations by defining new arithmetic rules on real numbers.

\subsubsection{Complex Number Representation}

Assuming we have a complex number $c=a+bi$, where $a$ is the real part and $bi$ is the imaginary part, we should use 2 real-valued channels to receive them. For example, assuming that the input has $N_{in}$ complex numbers, the kernel size is $m\times m$ and the output has $N_{out}$ feature maps, then we have a complex tensor $\bm{T}$, with its size defined as
\begin{equation}
\label{Eq:cplx_tensor}
Size_{\bm{T}}=N_in\times N_out\times m\times m\times 2.
\end{equation}

\subsubsection{Complex Convolution}
In order to process complex tensor, we need to define complex convolution. A complex convolutional filter is determined by its weight matrix, which is expressed in the same way as a complex tensor $\bm{T}$, i.e., using 2 real-valued channels to receive real part and imaginary part. We define a complex weight matrix as follows:
\begin{equation}
\bm{W} = \bm{A} + i\bm{B},
\end{equation}
where $\bm{A}$ and $\bm{B}$ are real number matrices. Assuming that the input is a complex vector $\bm{h}=\bm{x}+i\bm{y}$ and $*$ is the convolution operator, to make complex convolution equivalent to a real-valued 2D convolution, we have
\begin{equation}
\label{Eq:cplx_conv}
\bm{W}*\bm{h} = (\bm{A}*\bm{x}-\bm{B}*\bm{y}) + i(\bm{B}*\bm{x}+\bm{A}*\bm{y}).
\end{equation}
In Eq.~(\ref{Eq:cplx_conv}), according to the complex number arithmetic rules, we calculate the tensor real part and imaginary part separately on 2 channels. The size of output tensor follows Eq.~(\ref{Eq:cplx_tensor}).

\subsubsection{Complex-valued Activations}
To process the complex-valued representations, several activation functions are proposed. In our experiment, we use modReLU~\cite{arjovsky2016unitary}
\begin{equation}
\begin{split}
{\rm modReLU}(z) &= {\rm ReLU}(|z|+b)e^{i\theta_z}  \\
&=
\begin{cases}
 (|z|+b)\frac{z}{|z|} \ &{\rm if} \ |z|+b \leqslant 0, \\
 0   &{\rm otherwise},
\end{cases}
\end{split}
\end{equation}
where $z\in\mathbb{C}$, $\theta_z$ is the phase of $z$, and $b\in\mathbb{R}$ is a learned parameter. ReLU prohibits neurons with output of less than 0. In the complex domain, modReLU uses the modulus instead. If $z$ is within the circle with radius $b$ around the origin 0, it is set to 0. Otherwise, it is reserved. Thus, the phase $\theta_z$, which is an important feature, can be represented.

As commonly used in the finite element method (FEM) analysis of the MIT problem, we discretize the sensing area using 512 triangles~\cite{hollaus2004fast}. FEM decomposes the continuous field into multiple sub-regions. Instead of solving the entire continuous field, using sub-regions can simplify the differential equations and provide approximate numerical solutions. To make a fair comparison, we apply the same discretization in our deep learning model. Thus, the output of our model is a real-valued $1\times512$ vector, with each entry representing a triangle area and the distribution map is reconstructed. The triangular map generated by the corresponding ground truth image is used as the training label.

Using the existing complex CNN components~\cite{trabelsi2018deep} and the classic network structure of U-net~\cite{ronneberger2015u}, we design our complex classification network. The input is first converted to complex-valued features and then sent to layers of complex convolution and pooling. Since the output resolution is higher than the input, up-sampling is required. We use skip-connection to concatenate the low-level features and high-level features at the same scale, then up-sample the extracted features. The final layer converts the complex-valued numbers to real-valued ones. Finally, we apply a dense connected layer to obtain the real-valued $1\times512$ vector that matches the triangulation field. We use binary cross-entropy (BCE) loss to train the model, which is shown as follows.
\begin{equation}
\label{eq:BCE}
loss_{ccnn}(o,t)= \frac{1}{n}\sum_iw_i(t_ilog(o_i)+(1-t_i)log(1-o_i)),
\end{equation}
where $o$ is the output vector, $t$ is the ground truth vector from real-world measurements, and $w$ is the label weights. The conductivities of the object and the surrounding area are assumed uniform in our setup.

\subsection{GAN Based Network}

Since the conductivity distribution field is discretized, the output of complex CNN is coarse-grained. To improve the image quality, we employ cGAN~\cite{mirza2014conditional} to enhance the output. Our model is based on the pixel2pixel design~\cite{isola2017image}.

Normally GAN contains a generator and a discriminator, which is described by the following objective function.
\begin{equation}
L_{GAN}(G,D)=\mathbb{E}_{x}[\log D(x)] + \mathbb{E}_{x}[\log(1-D(G(z)))],
\end{equation}
where $D$ maximizes $\log D(x)$ while $G$ minimizes $\log(1-D(G(z)))$, $x$ is the label image, and $z$ is random noise. Generator output $G(z)$ is a candidate image with the probability distribution of $x$ that maps from the prior distribution of $z$. Discriminator output $D(x)$ or $D(G(z))$ is a scalar scoring how close the input is, which is defined as the possibility of input belonging to $x$.

Conditional GAN is used to reconstruct the field image. It adds extra information $y$ to penalize generator outputs that are beyond the given information. The extra information $y$ can be labels or any other limitations. Here $y$ is set as the complex CNN results, forcing the GAN output to be more accurate. In practice, $y$ is fed into both discriminator and generator as an additional input layer. The objective function can then be rewritten as
\begin{equation}
\begin{split}
L_{cGAN}(G,D)= & \mathbb{E}_{x,y}[\log D(x|y)] + \\
& \mathbb{E}_{x,y}[\log(1-D(G(z|y)))].
\end{split}
\end{equation}

Moreover, the L2 distance of the output is also penalized using the ground truth data while the discriminator remains unchanged~\cite{pathak2016context}. After adding the penalty, our objective function becomes
\begin{equation}
L^*=\arg \min_G\max_D L_{cGAN}(G,D)+\lambda L_{l2}(G).
\end{equation}

A normal generator transforms a white Gaussian noise or random dropout~\cite{isola2017image} to the target distribution. Noise $z$ represents the latent space of data. However, in the MIT inverse problem, it requires a stable output. Thus, our generator only takes the images as input instead of random noise or random dropout. In this work, we also employ the encoder-decoder structure~\cite{hinton2006reducing}.

In general, we use 10 convolution layers for feature extraction. After every convolution layer, batch normalization and ReLU activation are installed. There is a max-pooling layer for every 2 convolution layers. The up-sampling consists of 4 transpose convolution layers that concatenate with symmetrical low-level features in the structure. The final output resolution is 256$\times$256.

The discriminator employs the design of PatchGAN~\cite{isola2017image}. In a regular GAN, the discriminator typically outputs a scalar representing the corresponding confidence. PatchGAN outputs a $N\times N$ matrix whose element means a patch of the input is fake or not. It is achieved by sending concatenated image pairs into a series of convolution layers. A patch is a small part of the input image. The patches represent different perspective fields of the discriminator that are sensitive to image details at different textures. In the MIT problem, this loss penalizes generator output to make it more distinguishable between different conductivity objects. Moreover, the discriminator loss corresponds to the high-frequency part of the input images. For the low-frequency part, we use $L1$ loss to restrict the generated images.

\subsection{The Algorithms}
The training processes of complex CNN and GAN are shown in Algorithm~\ref{alg:cplxtrain} and Algorithm~\ref{alg:gantrain}, respectively. Compared with the original algorithms of the complex CNN and GAN, our algorithm has made several distinct modifications. First of all, we replace the original loss function using the BCE loss as shown in Eq.~(\ref{eq:BCE}). Since the output vector is binarized to reflect the existence of the object, the reconstruction process is actually a binary classification problem. Therefore, we use the BCE loss function, which does not require softmax to normalize the forward propagation. Moreover, to generate a stable output image, we initialize the GAN with the preprocessed images. Before training the complex CNN, we preprocess the data to get the input differential frame and ground truth map vector.

\begin{algorithm}[!h]
    \renewcommand{\algorithmicrequire}{\textbf{Input:}}
	\renewcommand{\algorithmicensure}{\textbf{Output:}}
    \caption{Training complex CNN}
    \label{alg:cplxtrain}
    \begin{algorithmic}[1]
        \REQUIRE
        differential measurement frames $\textbf{d}=\{d^1,d^2,...,d^n\}$,
        ground truth triangulation map vectors $\textbf{y}=\{y^1,y^2,...,y^n\}$.
        \ENSURE triangulation maps $\{\widetilde{x}^1|y^1,\widetilde{x}^2|y^2,...\}$.
        \REPEAT
        \STATE Sample $m$ examples $\{d^1|y^1,d^2|y^2,...,d^m|y^m\}$ from $\textbf{d}$ and $\textbf{y}$, correspondingly.
        \STATE Input examples into the complex CNN model and calculate the forward propagation result $\widetilde{y}^1,\widetilde{y}^2,...\}$
        \STATE Update complex CNN model parameters $\theta_{ccnn}$ to minimize:\\
        $loss_{ccnn}= \frac{1}{n}\sum_iw_i(y_ilog(\widetilde{y_i})+(1-y_i)log(1-\widetilde{y_i}))$.\\
        $\theta_{ccnn}\leftarrow\theta_{ccnn}+\eta\nabla loss_{ccnn}(\theta_{ccnn})$.
        \UNTIL convergence.
        \STATE Input differential measurement frame $d$ into the complex CNN model.
        \STATE Obtain triangulation map vectors $\{\widetilde{d}^1|y^1,\widetilde{d}^2|y^2,...\}$.
        \STATE Fill triangulation maps $\{\widetilde{x}^1|y^1,\widetilde{x}^2|y^2,...\}$ according to triangulation map vectors $\{\widetilde{d}^1|y^1,\widetilde{d}^2|y^2,...\}$.
        \STATE Return $\{\widetilde{x}^1|y^1,\widetilde{x}^2|y^2,...\}$.
    \end{algorithmic}
\end{algorithm}

\begin{algorithm}[!h]
    \renewcommand{\algorithmicrequire}{\textbf{Input:}}
	\renewcommand{\algorithmicensure}{\textbf{Output:}}
    \caption{Training GAN}
    \label{alg:gantrain}
    \begin{algorithmic}[1]
        \REQUIRE
        triangulation maps $\textbf{x}=\{x^1,x^2,...,x^n\}$,
        ground truth images $\textbf{t}=\{t^1,t^2,...,t^n\}$.
        \ENSURE reconstructed images $\{\widetilde{x}^1|t^1,\widetilde{x}^2|t^2,...\}$
        \REPEAT
        \STATE Sample $m$ examples $\{x^1|t^1,x^2|t^2,...,x^m|t^m\}$ from $\textbf{x}$ and $\textbf{t}$, correspondingly.
        \STATE Obtain $\{z^1|t^1,z^2|t^2,...,z^m|t^m\}$ by adding labels to the generated images.
        \STATE Obtain reconstruct images $\{\widetilde{x}^1|t^1,\widetilde{x}^2|t^2,...,\widetilde{x}^m|t^m\}$, where $\widetilde{x}^i|t^i=G(z|t^i)|t^i$.
        \STATE Update discriminator parameters $\theta_d$ to maximize:\\
        $v=\frac{1}{m}\sum_{i=1}^m \log D(x^i|t^i)+\frac{1}{m}\sum_{i=1}^m \log (1-D(\widetilde{x}^i|t^i))$.\\
        $\theta_d\leftarrow\theta_d+\eta\nabla v(\theta_d)$.
        \STATE Obtain $\{z^1|t^1,z^2|t^2,...,z^m|t^m\}$ by adding labels to the generated images.
        \STATE Update generator parameters $\theta_g$ to minimize:\\
        $v=\frac{1}{m}\sum_{i=1}^m \log [1-D(G(z|t^i)|y^i)] + \lambda\sqrt{\sum(\widetilde{x}^i-t^i)^2}$.\\
        $\theta_g\leftarrow\theta_g-\eta\nabla v(\theta_g)$.
        \UNTIL convergence.
        \STATE Obtain generated images $\{\widetilde{x}^1|t^1,\widetilde{x}^2|t^2,...\}$ by putting labels to generator: $\widetilde{x}^i|t^i=G(z|t^i)|t^i$.
        \STATE Return $\{\widetilde{x}^1|t^1,\widetilde{x}^2|t^2,...\}$.
    \end{algorithmic}
\end{algorithm}

\section{Experimental Results}
\label{sec:experiments}
In this section, we evaluate our algorithms based on the collected real-world dataset.

\subsection{Evaluation Metrics}

Intersection over union (IoU) is a common indicator for the evaluation of reconstructed images, which is defined as:
\begin{equation}
IoU=\frac{\rm{area}(R_{obj})\cap\rm{area}(G_{obj})}{\rm{area}(R_{obj})\cup\rm{area}(G_{obj})}\times100\%,
\end{equation}
where $R_{obj}$ is the object in the reconstructed image, $G_{obj}$ is the object in the ground truth map. A larger IoU indicates a better reconstruction of the image. When IoU equals 1, it means the reconstructed image is fully consistent with the ground truth.

Another indicator is the centroid distance (CD), which reflects the accuracy of locating the object. The centroid represents the mass center of a geometric object. Centroid distance is the Euclidean distance between the ground truth centroid and the reconstructed image centroid. It is defined as follows.
\begin{equation}
\label{Eq:cd}
CD=\sqrt{(x^*-x)^2+(y^*-y)^2},
\end{equation}
where $x^*, y^*$ is the centroid of ground truth and  $x, y$ is the centroid of reconstructed image. In Eq.~(\ref{Eq:cd}), $x$ and $y$ are calculated using:
\begin{equation}
x = \frac{1}{k} \sum^{k}_{i=1}x_i \text{ and } y = \frac{1}{k} \sum^{k}_{i=1}y_i.
\end{equation}

The CD between ground truth and reconstructed image should be as small as possible, while zero value means the two centroid are overlapped with each other.

\subsection{Baseline Methods}

We compare our MITNet with the Newton-Raphson (NR) method~\cite{wang2007image}, optimized fully-connected network (FCN)~\cite{xiao2018deep}, and stacked auto encoder (SAE)~\cite{li2019novel}.

\subsubsection{NR}

it solves the MIT problem based on the triangulation map. The variational equation representing the object field is~\cite{yorkey1987comparing}:
\begin{equation}
\begin{split}
F(\bm{A}) = & \frac{1}{2}\int \limits_{\Omega}[\mu^{-1}(x,y)(\frac{\partial^2 \bm{A}}{\partial x^2} + \frac{\partial^2 \bm{A}}{\partial y^2}) + \\
 & j\omega \sigma(x,y)\bm{A}^2]\mathrm{d}x\mathrm{d}y - 2\oint \limits_{L} \bm{A}J_{0}\mathrm{d}l,
\end{split}
\end{equation}

\begin{equation}
\frac{\partial F}{\partial \bm{A}} = 0,
\label{NRequation}
\end{equation}
where $\Omega$ is the measurement area, $L$ is the boundary of sensing coil, and $J_{0}$ is the excitation current density. By solving Eq.~(\ref{NRequation}), we can derive the voltage of sensing coil:
\begin{equation}
U = -j\omega \oint \limits_{l}\bm{A}\mathrm{d}l.
\end{equation}
As a result, the conductivity distribution $\bm{\sigma}$ is discretized into a 512$\times$ 1 matrix.

\subsubsection{FCN}

since FCN~\cite{xiao2018deep} is real-valued, we use the magnitude of complex-valued data as input. The output is the same as the complex CNN and NR algorithm. The input is a $1\times256$ vector, and the output is $1\times512$. The whole network contains 2 hidden layers of size $1\times360$, and there are batch normalizations between each layer. We fill the triangulation map with the corresponding output value.


\subsubsection{SAE}

It uses a stacked structure with 2 layers of autoencoders~\cite{li2019novel}. We set the 2 autoencoders output layers 512, input layer 256 and 128, and hidden layers 128 and 64, respectively. The two autoencoders are first trained using $1\times512$ ground truth vectors and the previous autoencoder hidden layer is used as the input of the next one. The final stacked autoencoder is consist of 2 layers of autoencoders together with a decoder layer of output size 512. The pre-trained autoencoders fit the structure of the training data, which makes the initial value of the entire network in a suitable state and speeding up training convergence.

\begin{figure*} [t]
\centering
\includegraphics[width=1.0\textwidth]{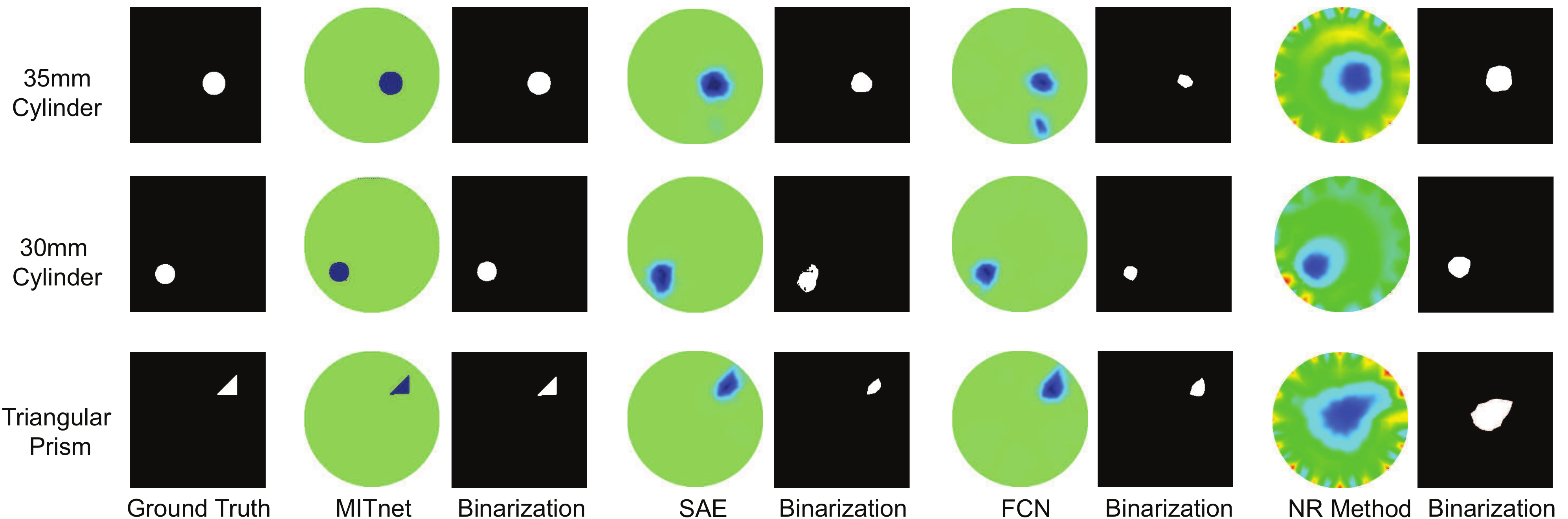}
\caption{The reconstruction results of all methods. Note that we calculate IoU and CD after binarization.}
\label{Fig:all_result_dem}
\end{figure*}

\subsection{Results and Analysis}

Our model is trained on a Tesla V100, Intel Xeon \SI{2.30}{\giga\hertz} CPU server with PyTorch. All networks use Adam optimizer with 0.001 learning rate, the training epoch is 200 and the batch size is set to 16.

After discretization, we smooth the data by averaging the value of each triangular region with the adjacent ones using the following equation.
\begin{equation}
\sigma_i=\frac{1}{m}\sum_{e=1}^m\sigma_i^e,
\end{equation}
where $m$ is the number of surrounding regions. This smoothing process is applied to all the considered algorithms.

Fig.~\ref{Fig:all_result_dem} shows the reconstruction images of all three kinds of data. The results demonstrate that our MITNet matches the ground truth best in all cases. It can restore the shape and location from the measured signal most accurately. By comparison, the NR method has large noises, while FCN and SAE are not able to reconstruct the object shape correctly.


\begin{figure} [!t]
\centering
\includegraphics[width=0.45\textwidth]{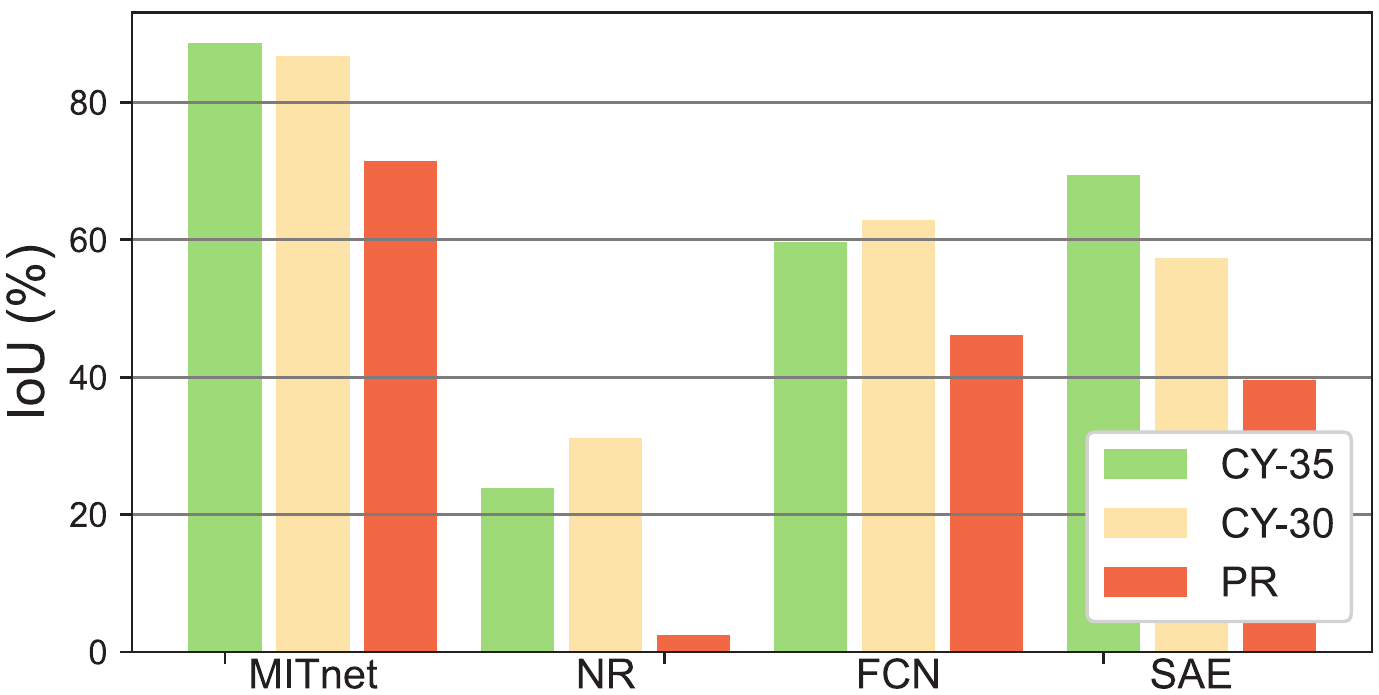}
\caption{IoU on all data.}
\label{Fig:iou}
\end{figure}

\begin{figure} [!t]
\centering
\includegraphics[width=0.45\textwidth]{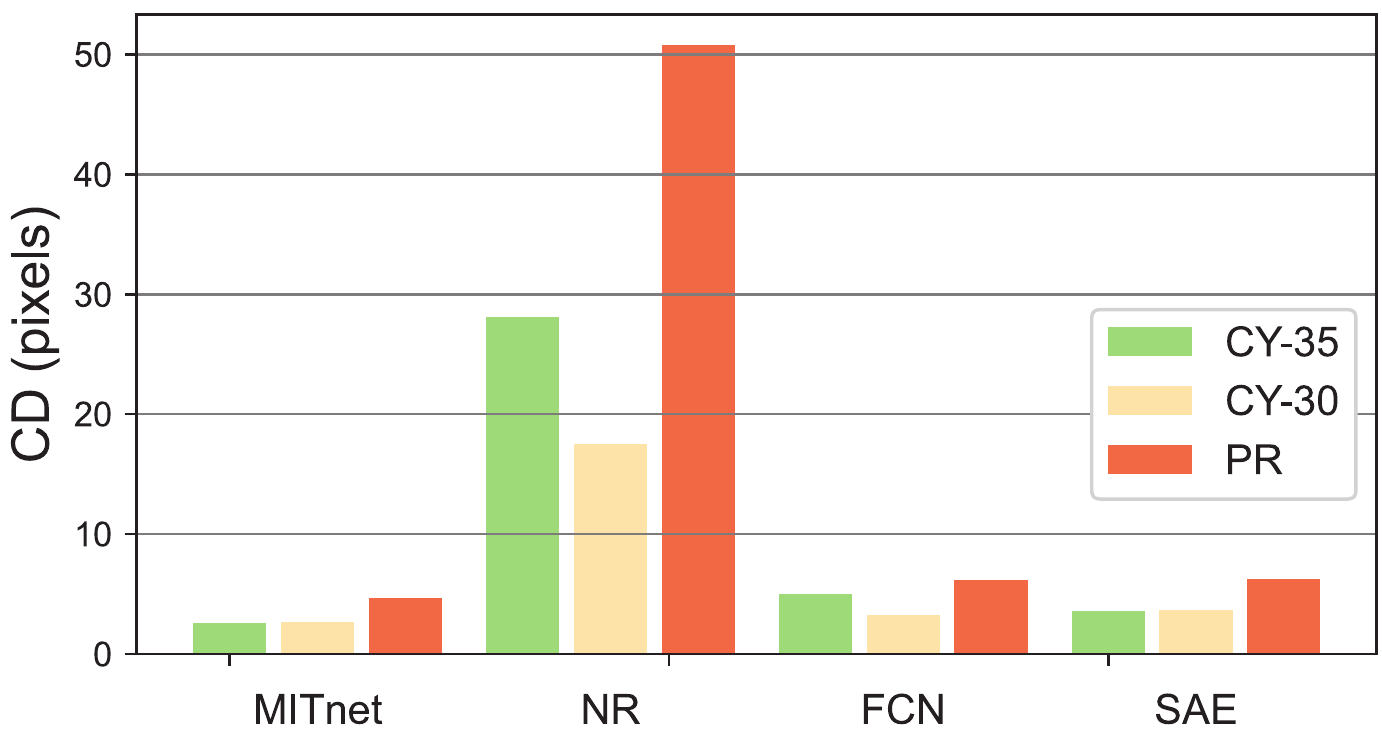}
\caption{CD on all data.}
\label{Fig:cd}
\end{figure}

As shown in Fig~\ref{Fig:iou} and Fig~\ref{Fig:cd}, MITNet achieves the best performance with highest IoU (82.25\%) and lowest CD (3.31) on average. It generates the reconstructed image closest to the ground truth. The reason is that our MITNet model takes the full information of the complex-valued signals. The NR method uses approximation during the calculation and can be easily affected by noise. The SAE and FCN take only the magnitude of data as input and thus, there is information loss in the training and testing process.

When the diameter of object changes, e.g., changing from \SI{35}{\milli\meter} to \SI{30}{\milli\meter}, the MITNet can reconstruct the images accordingly, achieving IoU of 88.61\% and 86.68\%, respectively.

On the triangular prism data, MITNet also generates better images than other algorithms. It has the highest IoU of 71.47\%. Since through finite element division, the resolution of origin output becomes coarse-grained, even after smoothing. It is very hard to recognize the shape and size of the object. For example, in Fig.~\ref{Fig:all_result_dem}, the reconstructed images of SAE on cylinder and prism are very close, but MITNet can identify the prism images with sharp edges.

\begin{table}[!t]
	\caption{IoU and CD of using GAN for all methods (Unit: \%/pixels).}
	\label{Tab:iou_cd_gan}
    \centering
	\begin{tabular}{|p{1.2cm}|>{\centering}p{1.2cm}>{\centering}p{1.2cm}>{\centering}p{1.2cm}>{\centering}p{1.2cm}|}
		\hline
		 IoU/CD     & CY-35 & CY-30 & PR & Average \tabularnewline
        \hline
	    NR          & 90.12/2.47	& 85.05/3.35 & 23.61/95.32 & 66.26/33.71 \tabularnewline
		\hline
        FCN         & 84.78/3.19	& 85.95/2.90 & 74.20/4.32 & 81.64/3.47 \tabularnewline
        \hline
        SAE         & 86.46/3.63	& 82.48/3.93 & 69.70/4.83 & 79.55/4.13 \tabularnewline
        \hline
        MITNet      & 88.61/2.62	& 86.68/2.63 & 71.47/4.68 & \textbf{82.25/3.31} \tabularnewline
        \hline
	\end{tabular}
\end{table}

To better understand the nature of our algorithm, we enhance the output of NR, FCN, and SAE algorithms using the same GAN technique. The results are listed in TABLE~\ref{Tab:iou_cd_gan}. After enhancement, all methods show improvement on different data. SAE and NR method shows great performance on both \SI{35}{\milli\metre} and \SI{30}{\milli\metre} cylinder data, which obtain IoU 90.12\%/85.05\% and 86.46\%/82.48\%, respectively. The CD is also lower after the enhancement. However, they have poor performance on the triangular prism data, which means that they have a limited ability to recognize object shapes. The results validate the effect of GAN in MIT image reconstruction. Overall, MITNet is still the best algorithm on all the data, achieving the highest IoU and the lowest CD on average.

\section{Conclusion}
\label{sec:conclusion}
In this paper, we design a bio-impedance distribution imaging system and propose a novel deep learning framework, MITNet, for image reconstruction. The system uses a set of excitation-sensing coils, inducing eddy current in the internal object field with a fixed frequency current, which is radiation-free and has potential for application in monitoring and diagnosis of cerebrovascular diseases. We also collect a real-world dataset based on our machine and evaluate our method together with existing algorithms. The experimental results show that the images reconstructed by our MITNet gain the highest quality, indicating that our system has great potentials for future biomedical imaging studies.


\end{document}